\documentclass[journal]{IEEEtran}

\usepackage{graphicx}
\usepackage[numbers,sort&compress]{natbib}
\usepackage{floatpag}
\usepackage{multirow}
\usepackage{amssymb}
\usepackage{color}
\usepackage{changepage}
\usepackage{array}

\graphicspath{{figures/}}

\begin{document}

\title{Knowledge transfer for surgical activity prediction}

\author{Olga Dergachyova, Xavier Morandi, Pierre Jannin
\thanks{Olga Dergachyova, Xavier Morandi and Pierre Jannin are affiliated with INSERM, U1099, Rennes, France and Universit\'e de Rennes 1, LTSI, Rennes, France (email: pierre.jannin@univ-rennes1.fr)}
\thanks{Xavier Morandi is affiliated with Service de Neurochirurgie, CHU Rennes, Rennes, France.}
}%

\maketitle

\begin{abstract}
Lack of training data hinders automatic recognition and prediction of surgical activities necessary for situation-aware operating rooms. We propose using knowledge transfer to compensate for data deficit and improve prediction. We used two approaches to extract and transfer surgical process knowledge. First, we encoded semantic information about surgical terms using word embedding which boosted learning process. Secondly, we passed knowledge between different clinical datasets of neurosurgical procedures using transfer learning. Transfer learning was shown to be  more effective than a simple combination of data, especially for less similar procedures. The combination of two methods provided 22\% improvement of activity prediction. We also made several pertinent observations about surgical practices.
\end{abstract}


\section{Introduction}
Automatic situation-awareness in the operating room (OR), including recognition and prediction of surgical workflow, is crucial for optimization of surgical process and OR management, decision support, intra-operative assistance, training and objective assessment of surgeons. Realization of these applications require extensive amounts of data. Meanwhile, lack of training data is a well-known problem in surgical domain. Multiple constraints impede  data acquisition: ethical approvals, patient's and medical staff's consents, limited amount of cases, expensive installation of acquisition equipment and tedious time-consuming manual annotations requiring medical experience. \textit{Surgical activities}, describing an action performed by the surgeon at the moment, used surgical instrument, and operated anatomical structure, represent a particular difficulty. These activities have short duration, high diversity in terms of number and sequencing. Very few works studying automatic recognition and prediction of such surgical activities exist \cite{Lalys13,Meissner14,Forestier17}. 

Today, deep learning greatly overcomes classical machine learning methods in many fields. Unfortunately for surgical domain, quantity of training data becomes a major factor. Massive datasets used by deep learning approaches, such as famous ImageNet, recently released Google's Open Images and YouTube-8M, contain millions of samples representing thousands of categories. These datasets constantly continue to grow and new ones keep appearing regularly. 

In machine learning, an approach called \textit{knowledge transfer} exist to overcome the problem of small training datasets. It involves methods that use resources from other domains of interest (generally having more training samples) to improve learning of a targeted task. \textit{Transfer learning}, a technique of knowledge transfer, is now widely used together with Convolutional Neural Networks (CNN) for tasks related to visual content (e.g., image/video recognition, captioning, segmentation and detection) benefiting from freely available image datasets \cite{Oquab14,Karpathy14}. It is also broadly applied to tasks of sequence analysis as speech and language processing \cite{Huang13} and document classification \cite{Dai07}. In surgical domain, transfer learning within CNN has been already used by Shin et al. \cite{Shin16} for classification of lung diseases, and by Twinanda et al. \cite{Twinanda17} for classification of surgical phases in laparoscopic surgeries. These studies transfered visual information only. Up until now, no research on transfer of surgical process knowledge (i.e., knowledge about how a procedure is performed in terms of its workflow) has yet been published. Transfer of surgical process knowledge has several difficulties comparing to classic visual or text-based transfer. First of all, any image can easily be brought to a needed format (size, color channels, etc.), any text  can be represented as a sequence of individual words. Surgical procedures, however, have their own vocabularies of surgical terms with different sizes and content describing the workflow. Secondly, available data both for visual and text-related tasks exist in abundance. Yet, the absence of comprehensive datasets describing complex processes as surgery limits transfer into the domain of surgical process. 

This paper exposes two contributions. First of all, we performed transfer of surgical process knowledge using two methods in order to overcome the lack of data and we demonstrated their efficiency for surgical activity prediction. The first transfer method of word embedding served to extract semantic knowledge about surgical terms from medical texts. The second consisted of transfer learning that enabled capturing important information about the surgical process and transferring it from one surgery to another. We conducted multiple experiments on different transfer sources and types to find best configurations. To the best of our knowledge, this is the first time when the transfer of surgical process knowledge was performed. Our findings can serve as a guidance for dataset enlargement for multiple recognition and analysis tasks involving surgical workflow. The second contribution consists in pertinent observations concerning surgical practices that help better understand the surgical process.

\section{Methods}

\begin{table*}[!tp]
\setlength{\tabcolsep}{13pt}
\centering
\caption{General information about the datasets. L and R stand for Leipzig and Rennes}
\label{table:general_info}
\begin{tabular}{c|l|cc|cc|cc}
\hline\noalign{\smallskip} 
\multicolumn{2}{l|}{Dataset} & ACDF.L & ACDF.R & LDH.L & LDH.R & PA.L & PA.R \\
\noalign{\smallskip}
\hline
\noalign{\smallskip}
\parbox[t]{3mm}{\multirow{6}{*}{\rotatebox[origin=c]{90}{Number of}}} &
interventions	  & 16 	& 48 	& 25 	& 20 	& 15 	& 11 \\
& activ. per inter-n   & 367$\pm$149 & 244$\pm$76 & 242$\pm$72 & 148$\pm$49 & 266$\pm$77 & 213$\pm$46 \\
& unique activities & 377 & 379 & 413 & 243 & 282 & 255 \\
& verbs 			  & 11 & 11 & 11 & 10 & 14 & 14 \\
& instruments 	  & 24 & 31 & 26 & 22 & 30 & 29 \\
& structures 		  & 10 & 6 & 9 & 7 & 6 & 8 \\
\noalign{\smallskip}
\hline
\end{tabular}
\end{table*}

\begin{table*}[!tp]
\setlength{\tabcolsep}{16pt}
\centering
\caption{Proportion of activities and activity elements shared between surgical sites and procedures, in \%}
\label{table:common_info}
\begin{tabular}{l|c|c|c||c|c|c}
\hline\noalign{\smallskip}
& \multicolumn{3}{c||}{inter-site} & \multicolumn{3}{c}{inter-procedure} \\
\noalign{\smallskip}
\hline
\noalign{\smallskip}
& ACDF 	& LDH 		& PA 		& ACDF x LDH  	& ACDF x PA 		& LDH x PA \\
\noalign{\smallskip}
\hline
\noalign{\smallskip}
Activities & 31.5 & 37.8 & 35.0 & 30.3 & 1.5 & 2.7 \\
Verbs & 100 & 95.2 & 100 & 100 & 78.6 & 78.6 \\
Instruments & 68.8 & 65.1 & 73.5 & 65.7 & 60.5 & 57.9 \\
Structures & 60 & 77.8 & 100 & 50 & 15 & 15 \\
\noalign{\smallskip}
\hline
\end{tabular}
\end{table*}

\subsection{Predicting next surgical activity}
\label{sec:prediction}
Automatic prediction of following surgical activities from already performed ones requires the system to deeply understand the surgical process of a given procedure. That is why this task represents a great challenge and at the same time a good example to demonstrate the efficiency of knowledge transfer. In this work, a surgical activity $A$ was defined as a 6-tuple $<L(V,I,S),~ R(V,I,S)>$  containing a verb $V$ describing the movement performed by the surgeon, instrument $I$ used for its execution, and operated anatomical structure $S$, separately specified for both left and right surgeon's hands. Let $\mathbb{P} = \{\alpha,\mathbb{S}\}$ describe the surgical process with a set of $m$ possible surgical activities $\alpha = \{A_1, A_2,..., A_m\}$ and a set $\mathbb{S}$ containing recorded surgical interventions represented as an ordered sequence of activity tuples $Seq = (a_1, a_2,..., a_l) \in \mathbb{S}$, where $a_i \in \alpha$, and the sequence length $l$ is different for each intervention. Let $Seq^*_t = (a_{t-n+1}, a_{t-n},..., a_t)$ define the workflow of an on-going intervention at current activity $t<l$ with a partial sequence of $n$ past activities. The learning task $\tau = \{\alpha, f(\cdot)\}$ consists in learning from training data an objective predictive function $f(Seq^*_t) = P(a_{t+1} = A_j | Seq^*_t)$ which predicts the next surgical activity for a sequence of already known performed activities. 

\begin{figure*}[h]
\vspace{-0.3cm}
\centering
\includegraphics[width=1\textwidth]{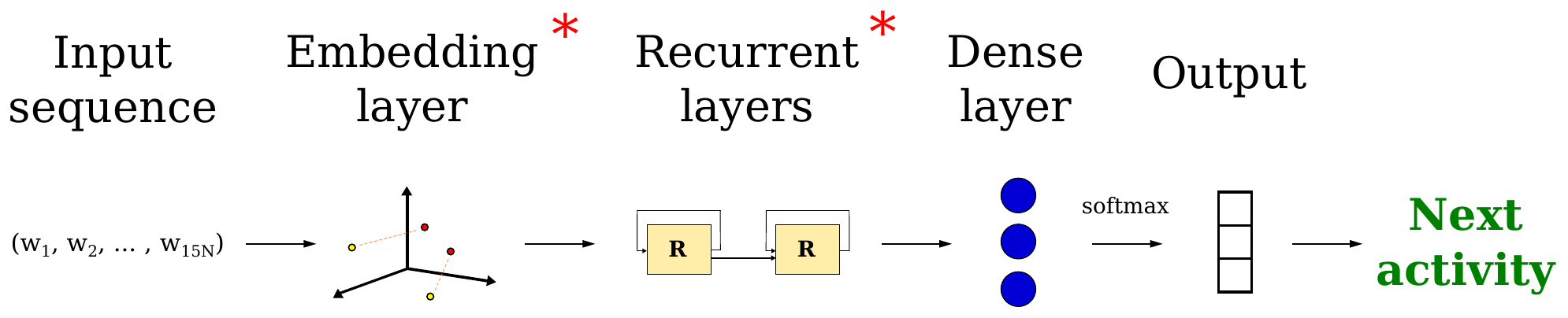}
\caption{Used LSTM model. Layers marked with a star are transferred}
\label{fig:model}
\end{figure*}

\subsection{Clinical data}
We studied the effect of knowledge transfer on three different neurosurgical procedures performed by junior and senior surgeons and acquired at university hospitals of Leipzig (Germany) and Rennes (France): anterior cervical discectomy and fusion (ACDF) \cite{Forestier13}, lumbar disc herniation (LDH) \cite{Riffaud10}, and pituitary adenoma (PA) \cite{Lalys10}. For each acquired intervention surgical workflow was annotated in terms of surgical activities defined as in Section \ref{sec:prediction} consisting of three activity elements (i.e., verb, instrument and structure) for each hand. An example of an ACDF activity could be \textit{(hold, forceps, disc, cut, scalpel, ligament)}. 
A total of 135 interventions were studied representing intra- and inter-procedure diversity, variety of practices, and a range of skill levels. Table \ref{table:general_info} contains additional information about each of the six datasets, and table \ref{table:common_info} shows the proportion of common activities and activity elements between both sites within one procedure and between two procedures from both sites.

\subsection{Word embedding}
\label{sec:word_embedding}

Usually, in deep learning approaches, words in a text-based dataset are represented as indices in a vocabulary. With such a representation, the words lose their semantics (i.e., no matter their meaning, they all become equally different from each other). To overcome this problem researchers in natural language processing (NLP) domain often use \textit{word embedding} technique that seeks to map semantic meaning of the words into a geometric space. This is done by associating a vector of real numbers to every word in the vocabulary so that the distance between the vectors captures semantic relationships between the corresponding words. In this work, we applied word embedding to extract, encode and re-inject into the model, the knowledge about the surgical domain and semantic meaning of surgical terms used to describe procedural workflow. We used two different methods of embedding: \textit{word2vec} (continuous bag-of-words configuration) \cite{Mikolov13} and \textit{GloVe} \cite{Pennington14}. Word2vec is a model training word embeddings via neural networks. It efficiently learns embedding vectors in order to improve the ability of predicting a target word from context words. The GloVe method learns embeddings by performing dimensionality reduction on the words co-occurrence matrix.

\subsubsection{Word corpora}

In order to obtain representative word embeddings, a \textit{word corpus} is necessary. It represents a collection of texts on a subject brought to the form of a plain sequence of words separated by single spaces. Various large corpora on general subjects are available on the Internet, as well as their pre-trained embeddings, yet there are very few corpora for medical or surgical fields. Therefore, we created three word corpora suitable for our purpose. Our first corpus Medical Transcription (MT), containing post-operative reports dictated by neurosurgeons, was  downloaded from iDASH repository (https://idash-data.ucsd.edu) which gathers the transcriptions from www.medicaltranscriptionsamples.com site. This corpus consisted of 103 transcriptions and contained 58975 in total and 4469 unique words. We composed the second corpus by automatically collecting the abstracts referenced by PubMed concerning three chosen neurosurgical procedures. The 62489 downloaded abstracts contained $>$57M words in total and 188K unique ones. We created the last corpus by collecting freely available full-text articles from PubMed Central (PMC) mentioning the same three procedures. The 32271 articles contained 118M words in total including 638K unique words. Jointly, these three corpora contained 708K unique words and 175M in total. Common vocabulary of clinical procedures studied in this work contains 79 unique words. Only 66 of them can be found in the MT corpus and all in PubMed and PMC corpora.

\subsection{Long Sort-Term Memory model for prediction}
\label{sec:lstm}

Long Short-Time Memory (LSTM) is a type of recurrent neural network usually applied for sequence-based problems where the items in an analyzed sequence have complex dependences. It has been shown to be very effective in many learning tasks. In this study, we performed the prediction of next surgical activity using LSTM to analyze the workflow of the procedure. A variant of classic LSTM including three gates (i.e., input, forget, output), an output activation function, no peephole connections, dropouts, and a full gradient training was used \cite{Graves12}. We also integrated word embeddings into the model to enable on-the-fly transformation of activity descriptions into meaningful representations. An embedding layer was put just after the input and before recurrent layers (Figure \ref{fig:model}). The values of this embedding layer are initialized from pre-trained embeddings importing only the words from the analyzed procedure. We tested different sets of LSTM parameters, each time varying the number of layers, number of hidden neurons, batch size, learning rate, optimizer, activation and loss functions, sequence size and different data representations. The best configuration had the following parameters. It contained two stacked recurrent layers with dropouts of 0.2, each having 256 hidden neurons. It was trained during 50 epochs with a learning rate of 0.001 by 128-size batches. Categorical cross entropy was used as the loss propagation function, together with Adam optimizer. The best results were also obtained by analyzing a period of 75 past activities for ACDF.L and LDH.L and 50 for all other datasets. Each input activity meaning its elements were transformed into a sequence of words that were then normalized (padded or truncated if needed) to have a common size of 15. Each analyzed period thus consisted of a concatenation of the word sequences of contained activities.

\vspace{-0.4cm}
\subsection{Transfer learning}
\label{sec:transfer}

In surgical process modeling, the problem of lack of training samples can be interpreted as a lack of knowledge about a surgical procedure and its execution. We applied transfer learning approach to this problem motivated by the following hypothesis. Sequences of activities representing a surgical process encode some form of abstract knowledge about a given procedure, surgical practice and the process in general. This knowledge can be extracted and exploited to improve all sorts of operations on surgical process data, including analysis, recognition and prediction. It was particularly assumed that the knowledge obtained from one procedure might improve prediction of surgical activities for another procedures. In view of the implicit nature of the knowledge that can hardly be formalized, deep recurrent neural networks are a good tool able to extract and transfer it. 

In this work we performed transfer learning to transfer surgical process knowledge between different clinical datasets. We called the dataset containing knowledge extracted for transfer \textit{source dataset} $D_{src}$, and the model extracting knowledge by training of the network on the source dataset \textit{source model} $M_{src}$. Analogically, the dataset that benefited from transferred knowledge was called \textit{destination dataset} $D_{dist}$, and the model that was initialized with the weights of the source model was called \textit{destination model} $M_{dist}$. The transfer between the source and destination models of this type is made as follows. For the source model $M_{src}$, the embedding layer is set from pre-trained embeddings, the recurrent and dense layers are initialized randomly. Then, the model (all layers) is trained on the source dataset $D_{src}$. In deep models, the knowledge learned from the data is encoded in the weight matrices representing the layers' internal parameters. The weight matrices from the embedding and both recurrent layers are exported and saved for further transfer. For the destination model $M_{dist}$, the embedding and both recurrent layers are first set with the weights from corresponding source layers. The dense layer is randomly initialized. The $M_{dist}$ model is then trained on the destination dataset $D_{dist}$ updating the weights of all layers. The weights of the dense layer can not be transfered as the number of its internal parameters is tied up with the number and content of activities in the source dataset. Despite a large fraction of common elements, the procedures still have small number of common activities. Put all possible activities from both source and destination procedures into the dense layer would be suboptimal. Withal, the main knowledge is supposed to be encoded in the recurrent layers, while the dense layer only serves to connect observations to the activities.

\subsection{Study design}
We performed several experiments to assess the effect of two methods of knowledge transfer on surgical activity prediction. The main factor observed trough the entire study was $\Delta$ - the amount of prediction improvement between different configurations that is the difference in prediction accuracy. Prediction accuracy score was computed as the total amount of correctly predicted activities to the number of test samples. An activity was assumed to be correctly predicted if all its elements of both hands were correctly predicted. For each dataset we performed cross-validation, each time averaging the results of three training runs for every fold. The statistical significance of the results compared to base line values was measured using two-tailed Wilcoxon rank-sum test (** for p$\le$0.01, * for p$\le$0.05, no star for p$>$0.05).

\subsubsection{Word embedding}
First of all, a base line was defined as the prediction accuracy of a model with parameters from Section \ref{sec:lstm} but without any form of transfer. Then, the impact of pre-trained embeddings containing semantic information was estimated. This time, we added an embedding layer to the model and tried different configurations: embeddings integration (``set'' - simple initialization of the layer with pre-trained weights or ``set + train'' - initialization and additional training on clinical data along with other layers), embedding method (word2vec vs. GloVe), word corpus (MT, PubMed, PMC or all combined), and embedding vector size (100, 300 or 500). 

\subsubsection{Transfer learning}
To be able to transfer knowledge from one dataset to another, the networks used for training must have the same number of internal parameters in transferred layers, meaning that the length of analyzed sequences (i.e., number of past activities taken into consideration when making prediction) have to be equal. Thus, we also computed a new embedding base line for all datasets using the same network hyper-parameters and the best embedding configuration to objectively measure the improvements in prediction obtained by transfer learning. Besides that, in this part of the study, two types of experiments were made as described below.

\paragraph{Mix.} 
This experiment was performed to found out how simply putting different datasets together changes the prediction accuracy compared to application of transfer learning technique. It is different from transfer learning, as the model training is made on all the data at once.
The data was mixed in two different ways: 1) all interventions from both sites within one procedure (i.e., L + R in ACDF, LDH and PA), and 2) different pairs of procedures including interventions from both sites (i.e., ACDF (L+R) + LDH (L+R), ACDF (L+R) + PA (L+R) and LDH (L+R) + PA (L+R)).

\paragraph{Transfer.}
We conducted this experiment to measure the improvement provided by the actual transfer learning.
The learning was performed as described in Section \ref{sec:transfer}. We performed two types of transfer using different combinations of source and destination datasets: inter-site and inter-procedure transfers. The inter-site transfer, which can also be considered as intra-procedure transfer, was performed to estimate the effectiveness of transfer between the interventions performed in different hospitals belonging to the same procedure. It involved the following dataset pairs: L $\rightarrow$ R and R $\rightarrow$ L for ACDF, LDH and PA (the source dataset is on the left of the arrow sign, and destination on its right). The inter-procedure transfer was performed to estimate the effectiveness of transfer between different procedures belonging to the same surgical specialty. It involved the pairs as ACDF $\leftrightarrow$ LDH, ACDF $\leftrightarrow$ PA, LDH $\leftrightarrow$ PA (the double-side arrow means that the transfer was made from the left (source) dataset to the right (destination), and the other way around).

\vspace{-0.2cm}
\subsection{Results}

\subsubsection{Word embedding}
\begin{figure}[!tp]
\centering
\includegraphics[width=0.5\textwidth]{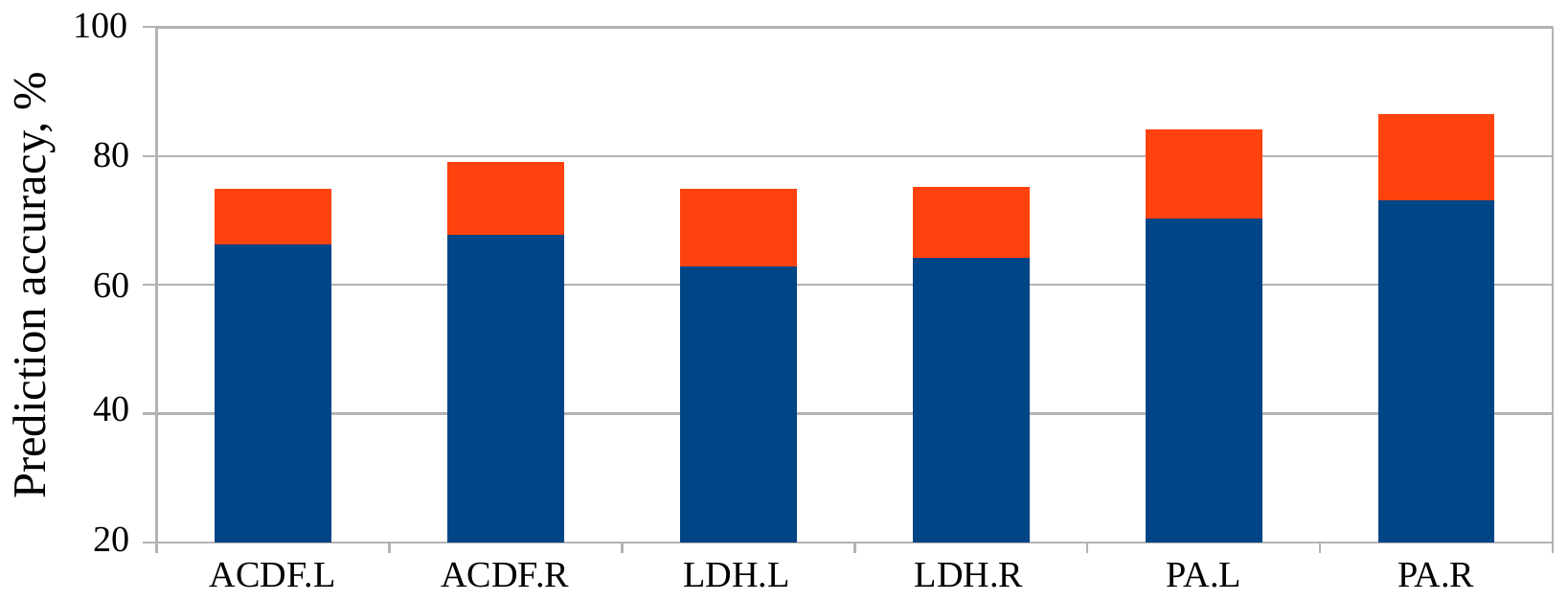}
\caption{Prediction accuracy (in \%) for different datasets for the base line (in blue) and best embedding configuration (red color indicates the improvement)}
\label{fig:best_embedding}
\end{figure}

The base line configuration (no embedding, no transfer learning) produced 67.4$\pm$12.5\% accuracy on average for all datasets. The ``set'' configuration (simple initialization of the embedding layer with pre-trained vectors) at best reached 75.7$\pm$9.1\%$^*$ ($\Delta$ = 8.3\%). These results were obtained with word2vec method by training of a 500-dimensional vector on conjunction of all corpora. The best results for ``set + train'' configuration, accuracy of 79.1$\pm$7.5\%$^{**}$ ($\Delta$ = 11.7\%), were achieved with the same parameters but using GloVe method. Figure \ref{fig:best_embedding} displays this improvement separately for all datasets. 

The PA activities were predicted the best and those belonging to LDH the worst. Better predictions were also made for the procedures performed in Rennes than Leipzig. 
We also observed that the accuracy of prediction generally went up with the growth of the corpus size. The same happened with the increase of the embedding vector size, except for the MT corpus (it seems to be too small for higher vector sizes). The word2vec method performed better than GloVe with smaller corpora, but both provided similar results for larger corpora. The pre-trained embeddings were made for a learning objective quite different form activity prediction. The additional training for the current objective adjusted the embeddings to reflect more the semantics of the surgical process and fairly improved the results. 

\subsubsection{Transfer learning}

The new embedding base-line model on average generated the accuracy of 78.9$\pm$7.5\%. This result is slightly lower than in the previous experiment because the chosen common sequence length of 50 activities provides the best results for only four of six datasets.

\begin{figure*}[h]
\centering
\includegraphics[width=1\textwidth]{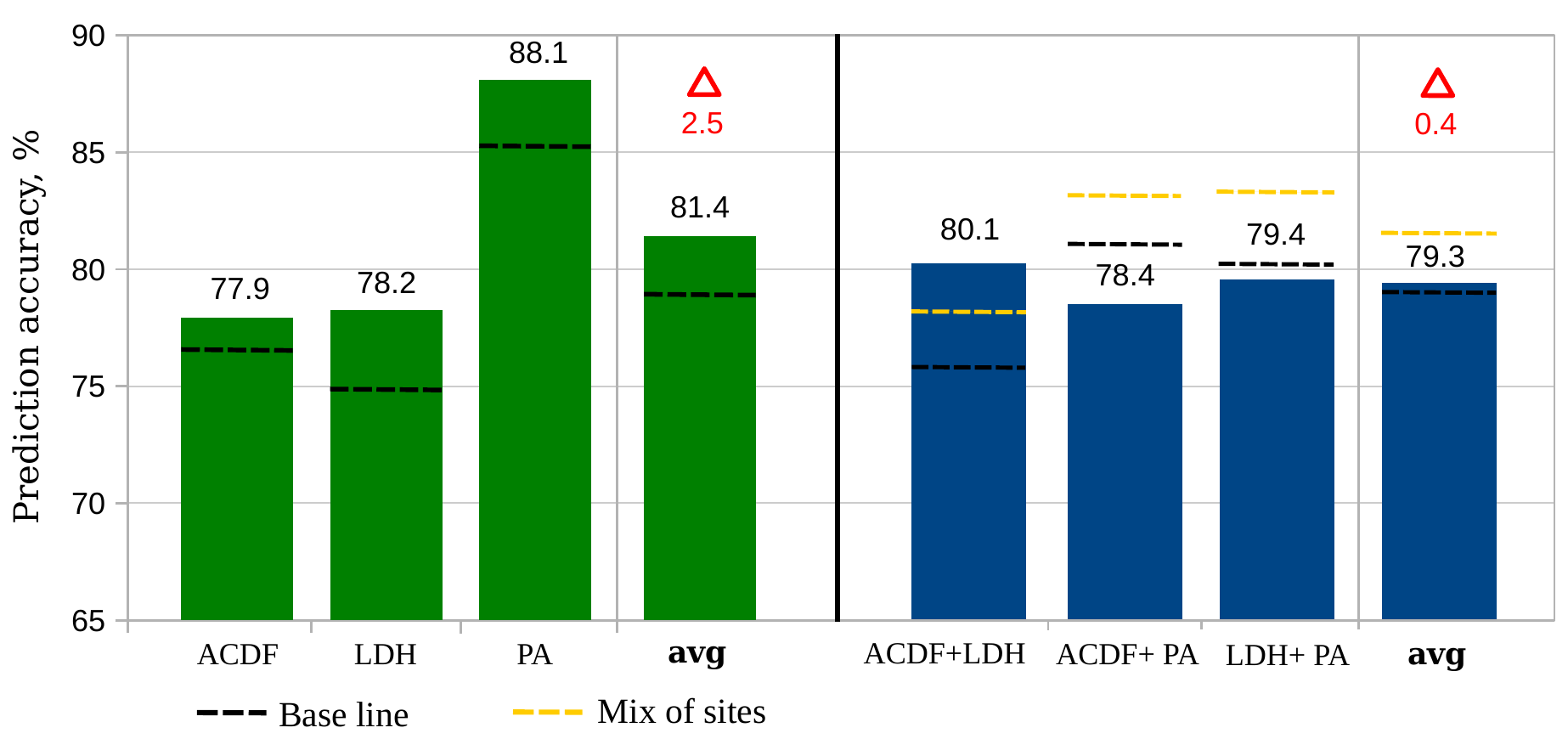}
\caption{Mix of datasets. The mix of sites within one procedure is on the left in green and the mix of different procedures from both sites on the right in blue}
\label{fig:mix}
\end{figure*}

\paragraph{Mix.} Figure \ref{fig:mix} shows the results of mix of datasets. Mixing the interventions belonging to the same procedure but performed at different surgical sites provided 81.4\%$^*$ accuracy ($\Delta$ = 2.5\% compared to the base line standard). LDH had the biggest increase in accuracy, while ACDF the lowest.
Mixing different procedures from both sites generated 79.3\% accuracy ($\Delta$ = 0.4\% compared to the base line standard). The combination ACDF + LDH had 4.98\%$^*$ gain over the base line, while the ACDF + PA lost 2.16\%$^*$, and LDH + PA lost 0.30\%. However, compared to the mix of sites within a procedure, the mix of procedures actually lost in accuracy with average $\Delta$ = -1.46\% (only the ACDF + LDH combination improved prediction performance of 3.62\%$^*$). 

\begin{figure*}[!bp]
\vspace{-0.3cm}
\centering
\includegraphics[width=1\textwidth]{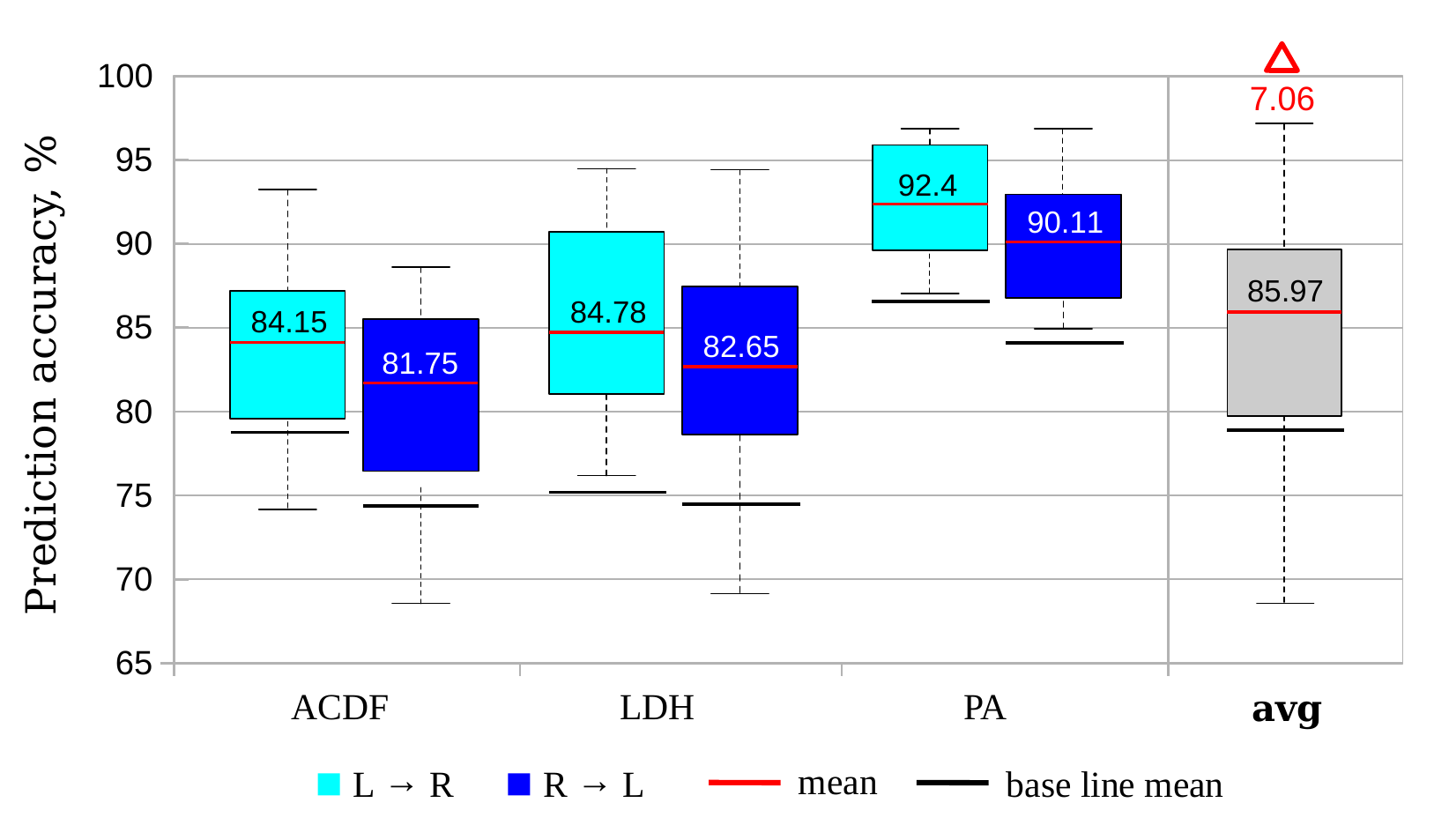}
\caption{Inter-site transfer}
\label{fig:transfer_site}
\includegraphics[width=1\textwidth]{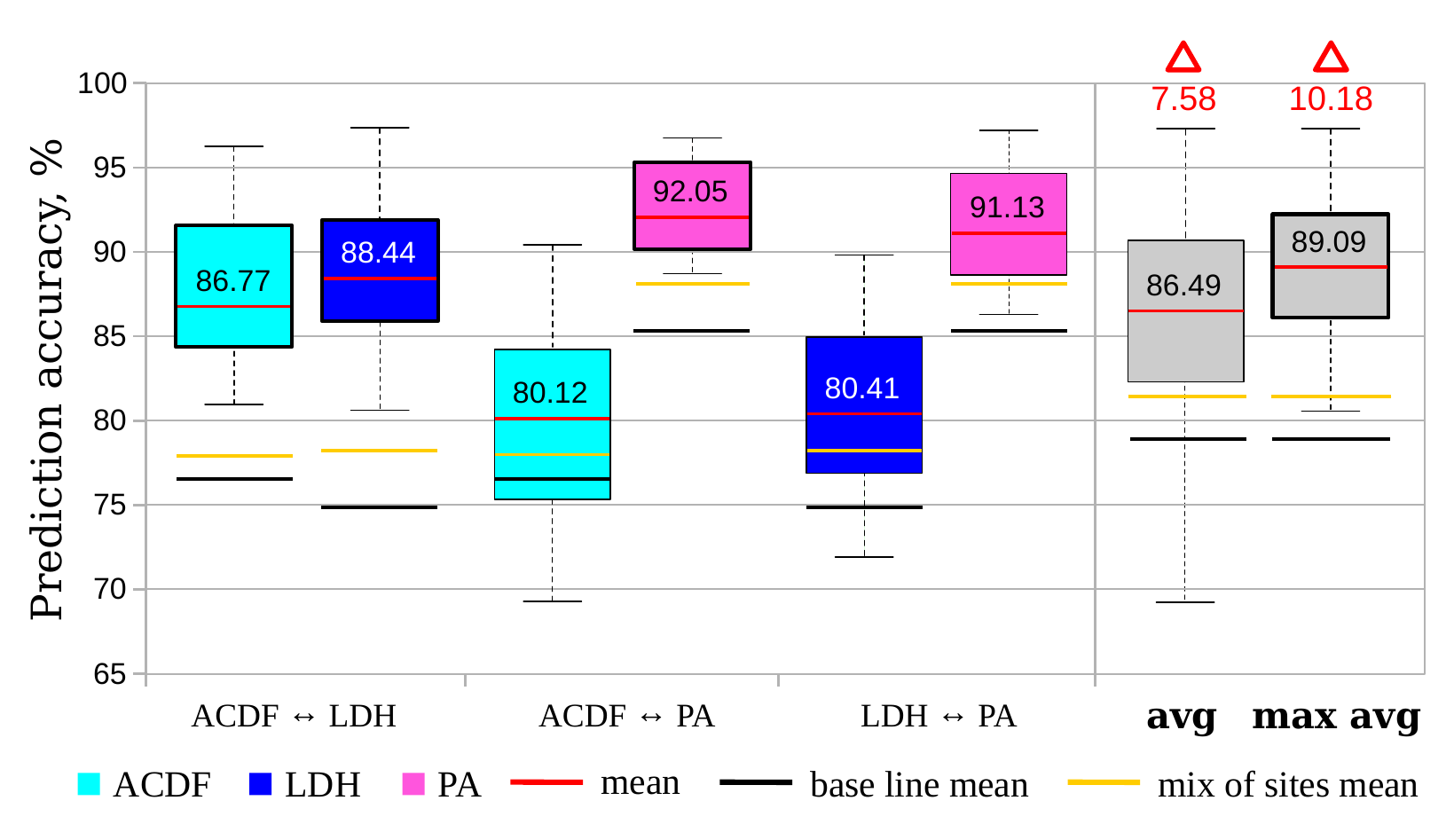}
\caption{Inter-procedure transfer}
\label{fig:transfer_surgery}
\vspace{-1.5cm}
\end{figure*}

\paragraph{Transfer.}
The results of the inter-site transfer are shown in Figure \ref{fig:transfer_site}. On average for all procedures and source-destination combinations, the prediction performance of the model achieved $\approx$86\%$^*$ ($\Delta$ = 7.1\% comparing to the base line). LDH benefited from the knowledge transfer the most, ACDF less but had a more important increase than PA. It shows that even if two sets of data are less similar, one can still have some form of knowledge (common or not) that enhances the learning process of the second one. The results of the inter-procedure transfer are exposed in Figure \ref{fig:transfer_surgery}. On average for all tested combinations of source and destination datasets, the prediction accuracy reached 86.5\%$^*$ ($\Delta$ = 7.6\% compared to the baseline). However, 2.5\% of this improvement came from the mix of both sites in the source and destination datasets. Thus, the actual increase in accuracy equaled 5.1\%. On the other hand, if we choose only the most appropriate procedures for transfer (e.g., ACDF for LDH, LDH for ACDF and ACDF for PA), the average accuracy can be recomputed to 89.1\%$^*$ with the $\Delta$ of 10.2\% for the base line and 7.7\% for the mix of sites. The highest improvement was made for LDH when transferring knowledge from ACDF, and the lowest for ACDF when transferring from PA. This experiment demonstrates, that even if the source and destination procedures are quite different from each other, both still encode some fundamental knowledge about the surgical process in general that can help in training. 

\section{Discussion}
The goal of our work was to achieve a significant improvement in activity prediction using knowledge transfer. The combination of both suggested for transfer methods increased the activity prediction ability of a basic LSTM model by almost 22\%. The best training configuration reached an accuracy of 89\% on average for all used datasets. This prediction performance was generated by a very basic LSTM architecture. Finding the best suitable architecture and hyper parameters was out of the scope of this work. However, a more elaborated model would probably provide better results. Moreover, the presented approaches of knowledge transfer can easily be applied to any surgical workflow related problem (e.g., recognition and analysis). 

We demonstrated that the word embeddings trained on corpora constructed from medical and scientific texts, indeed, contained certain semantic information that boosted learning process and played an important role in knowledge transfer. Relatively small word corpora were used for creating embedding vectors. Computed representation can be further improved with bigger corpora. Our corpora were based on neurosurgical post-operative reports and scientific articles. It would also be interesting to create and test a word corpus specifically dedicated to describe a surgical process.

For transfer learning, the two-step training process ``mix of sites + inter-procedure transfer'' produced the best improvement. The most effective transfer happened between the procedures that resembled the most: ACDF and LDH. However, the transfer between less related procedures, as ACDF $\leftrightarrow$ PA or LDH $\leftrightarrow$ PA, was also helpful. This showed that even having very few activities in common, two procedures can still contain certain fundamental knowledge or common information about the surgical process which can turn out to be very useful. In the future, a multi-step transfer has to be tested meaning learning by subsequently transferring information between more than two datasets. It is also possible to try importing the weights of different recurrent layers separately. The study results also allow us to think that the surgeries from other specialties or even non-surgical process-based sequences can be used for transfer as well. This opens the doors to many opportunities, as the acquisition of non-surgical process models is less constrained and may be used to create massive datasets to learn from.

\paragraph{Knowledge.} The knowledge extracted by LSTM networks probably contained both procedure-dependent (i.e., specific terms and concepts) and independent features (i.e., sequential information inherent to any surgical process). Unfortunately, a neural network, especially a recurrent, remains a black box. It is difficult to see what happens inside and to know how, why and what exactly it learns. Understanding of the learning process is an important research direction. Moreover, the extracted knowledge is also encoded a human-unreadable way; its is not formalized and, for now, can only be used within deep network architectures. Combining it with formal representations (e.g., ontology) would be highly relevant. 

\paragraph{Surgical practice.} During our study, we also made several observations about surgical practices. The obtained results suggested that the same procedures from different hospitals were more alike than different neurosurgical procedures from one hospital. However, the difference between two sites depended on the procedure. The LDH procedure was performed more similarly in Rennes and Leipzig than PA or ACDF. Another discovery demonstrated that the procedures in Rennes resembled more to each other than those in Leipzig, as probably the surgical process in Rennes was more standardized. 

\section{Conclusion}

In this work, focused on the problem of data deficiency, we proposed to use knowledge transfer methods in order to compensate small amounts of training data. To demonstrate their effect, the task of next surgical activity prediction was chosen as example. We proposed two transfer methods improving the prediction accuracy by almost 22\% in total. The first method was the word embedding technique vastly applied in natural language processing. It was used to extract semantic knowledge describing the relationships between surgical terms from specially created medical word corpora. The second transfer learning method passed the knowledge about the surgical process from one dataset of annotated interventions to another. The best results were obtained when transferring information from one procedure to another. Several pertinent observations about surgical practices were also made. This work is the first study in the literature applying knowledge transfer to surgical processes.

\section*{Acknowledgements}
This work was partially supported by French state funds managed by the ANR within the Investissements d'Avenir programme (Labex CAMI) under reference ANR-11-LABX-0004.

\bibliographystyle{IEEEtran} 
\bibliography{Transfer}

\begin{thebibliography}{10}
\providecommand{\url}[1]{#1}
\csname url@samestyle\endcsname
\providecommand{\newblock}{\relax}
\providecommand{\bibinfo}[2]{#2}
\providecommand{\BIBentrySTDinterwordspacing}{\spaceskip=0pt\relax}
\providecommand{\BIBentryALTinterwordstretchfactor}{4}
\providecommand{\BIBentryALTinterwordspacing}{\spaceskip=\fontdimen2\font plus
\BIBentryALTinterwordstretchfactor\fontdimen3\font minus
  \fontdimen4\font\relax}
\providecommand{\BIBforeignlanguage}[2]{{%
\expandafter\ifx\csname l@#1\endcsname\relax
\typeout{** WARNING: IEEEtran.bst: No hyphenation pattern has been}%
\typeout{** loaded for the language `#1'. Using the pattern for}%
\typeout{** the default language instead.}%
\else
\language=\csname l@#1\endcsname
\fi
#2}}
\providecommand{\BIBdecl}{\relax}
\BIBdecl

\bibitem{Lalys13}
F.~Lalys, D.~Bouget, L.~Riffaud, and P.~Jannin, ``Automatic knowledge-based
  recognition of low-level tasks in ophthalmological procedures,''
  \emph{International Journal of Computer Assisted Radiology and Surgery},
  vol.~8, no.~1, pp. 39--49, 2013.

\bibitem{Meissner14}
C.~Mei{\ss}ner, J.~Meixensberger, A.~Pretschner, and T.~Neumuth, ``Sensor-based
  surgical activity recognition in unconstrained environments,''
  \emph{Minimally Invasive Therapy \& Allied Technologies}, vol.~23, no.~4, pp.
  198--205, 2014.

\bibitem{Forestier17}
G.~Forestier, F.~Petitjean, L.~Riffaud, and P.~Jannin, ``Automatic matching of
  surgeries to predict surgeons' next actions,'' \emph{Artificial Intelligence
  in Medicine}, 2017.

\bibitem{Oquab14}
M.~Oquab, L.~Bottou, I.~Laptev, and J.~Sivic, ``Learning and transferring
  mid-level image representations using convolutional neural networks,'' in
  \emph{Proceedings of the IEEE conference on computer vision and pattern
  recognition}, 2014, pp. 1717--1724.

\bibitem{Karpathy14}
A.~Karpathy, G.~Toderici, S.~Shetty, T.~Leung, R.~Sukthankar, and L.~Fei-Fei,
  ``Large-scale video classification with convolutional neural networks,'' in
  \emph{Proceedings of the IEEE conference on Computer Vision and Pattern
  Recognition (CVPR)}, 2014, pp. 1725--1732.

\bibitem{Huang13}
J.-T. Huang, J.~Li, D.~Yu, L.~Deng, and Y.~Gong, ``Cross-language knowledge
  transfer using multilingual deep neural network with shared hidden layers,''
  in \emph{IEEE International Conference on Acoustics, Speech and Signal
  Processing (ICASSP)}, 2013, pp. 7304--7308.

\bibitem{Dai07}
W.~Dai, Q.~Yang, G.-R. Xue, and Y.~Yu, ``Boosting for transfer learning,'' in
  \emph{Proceedings of the 24th international conference on Machine learning},
  2007, pp. 193--200.

\bibitem{Shin16}
H.-C. Shin, H.~R. Roth, M.~Gao, L.~Lu, Z.~Xu, I.~Nogues, J.~Yao, D.~Mollura,
  and R.~M. Summers, ``Deep convolutional neural networks for computer-aided
  detection: Cnn architectures, dataset characteristics and transfer
  learning,'' \emph{IEEE Transactions on Medical Imaging}, vol.~35, no.~5, pp.
  1285--1298, 2016.

\bibitem{Twinanda17}
A.~P. Twinanda, S.~Shehata, D.~Mutter, J.~Marescaux, M.~de~Mathelin, and
  N.~Padoy, ``Endonet: A deep architecture for recognition tasks on
  laparoscopic videos,'' \emph{IEEE Transactions on Medical Imaging}, vol.~36,
  no.~1, pp. 86--97, 2017.

\bibitem{Forestier13}
G.~Forestier, F.~Lalys, L.~Riffaud, D.~L. Collins, J.~Meixensberger, S.~N.
  Wassef, T.~Neumuth, B.~Goulet, and P.~Jannin, ``Multi-site study of surgical
  practice in neurosurgery based on surgical process models,'' \emph{Journal of
  Biomedical Informatics}, vol.~46, no.~5, pp. 822--829, 2013.

\bibitem{Riffaud10}
L.~Riffaud, T.~Neumuth, X.~Morandi, C.~Trantakis, J.~Meixensberger, O.~Burgert,
  B.~Trelhu, and P.~Jannin, ``Recording of surgical processes: a study
  comparing senior and junior neurosurgeons during lumbar disc herniation
  surgery,'' \emph{Neurosurgery}, vol.~67, pp. 325--332, 2010.

\bibitem{Lalys10}
F.~Lalys, L.~Riffaud, X.~Morandi, and P.~Jannin, ``Automatic phases recognition
  in pituitary surgeries by microscope images classification,'' in
  \emph{International Conference on Information Processing in Computer-Assisted
  Interventions}, 2010, pp. 34--44.

\bibitem{Mikolov13}
T.~Mikolov, K.~Chen, G.~Corrado, and J.~Dean, ``Efficient estimation of word
  representations in vector space,'' \emph{arXiv preprint arXiv:1301.3781},
  2013.

\bibitem{Pennington14}
J.~Pennington, R.~Socher, and C.~D. Manning, ``Glove: Global vectors for word
  representation.'' in \emph{EMNLP}, vol.~14, 2014, pp. 1532--1543.

\bibitem{Graves12}
A.~Graves, \emph{Supervised Sequence Labelling with Recurrent Neural
  Networks}.\hskip 1em plus 0.5em minus 0.4em\relax Springer Berlin Heidelberg,
  2012.

\end{thebibliography}

\end{document}